\documentclass[12pt]{article}
\usepackage[utf8]{inputenc}
\usepackage[brazil]{babel}
\usepackage{sbc-template}
\usepackage{changepage}
\usepackage{graphicx}
\usepackage{wrapfig}
\usepackage{url}

\usepackage[
    justification=justified
]{caption}

\newcommand{\ReGENN}{{\sc ReGENN}}

\newcommand{\ie}{{\it i.e.}}

\title{From Cities to Series: Complex Networks and Deep Learning for Improved Spatial and Temporal Analytics%
\footnote{This piece refers to an extended abstract of the Ph.D. thesis under the same name defended in the Graduate Program in Computer Science and Computational Mathematics (PPG-CCMC) of the Institute of Mathematics and Statistics (ICMC) from the University of Sao Paulo (USP) -- Brazil on July 12, 2021. The authors here listed refer to the Ph.D. candidate and his advisor, respectively. This paper was generated while the first author was conducting postdoctoral studies at the Dalhousie Univerisity, Halifax -- NS, Canada.}}

\author{Gabriel Spadon\inst{1,2}, Jose F. Rodrigues-Jr\inst{1}}

\address{%
    Institute of Mathematics and Statistics (ICMC)\\
    University of Sao Paulo (USP), Sao Carlos -- SP, Brazil
    \nextinstitute%
    Institute for Big Data Analytics (IBDA)\\
    Dalhousie University (DAL), Halifax -- NS, Canada
    \email{gabriel@spadon.com.br, junio@icmc.usp.br}
}

\begin{document}

\maketitle

\begin{abstract}
The relationship between entities of interest is a property that can be represented as a graph defined over sets of entities (vertices) and relationships (edges).
Graphs have often been used to answer questions about the interaction between real-world entities by taking advantage of their capacity to represent complex topologies.
Complex networks are known to be graphs that capture such non-trivial topologies; they are able to represent human phenomena such as epidemic processes, the dynamics of populations, and the urbanization of cities.
The investigation of complex networks has been extrapolated to many fields of science, with particular emphasis on computing techniques, including artificial intelligence.
In such a case, the analysis of the interaction between entities of interest is transposed to the internal learning of algorithms, a paradigm whose investigation is able to expand the state of the art in Computer Science.
By exploring this paradigm, this thesis puts together complex networks and machine learning techniques to improve the understanding of the human phenomena observed in pandemics, pendular migration, and street networks.
Accordingly, we contribute with:
(i) a new neural network architecture capable of modeling dynamic processes observed in spatial and temporal data with applications in epidemics propagation, weather forecasting, and patient monitoring in intensive care units;
(ii) a machine-learning methodology for analyzing and predicting links in the scope of human mobility between all the cities of Brazil; and,
(iii) techniques for identifying inconsistencies in the urban planning of cities while tracking the most influential vertices, with applications over Brazilian and worldwide cities.
We obtained results sustained by sound evidence of advances to the state of the art in artificial intelligence, rigorous formalisms, and ample experimentation.
Our findings rely upon real-world applications in a range of domains, demonstrating the applicability of our methodologies.
\end{abstract}

\section*{Introduction}

The fusion of graph theory (and network science~\cite{Barabasi2016:NetworkScience}) with artificial neural networks (\ie, deep learning~\cite{Lecun2015:DeepLearning}) has revealed inspiring results in a myriad of domains~\cite{Zhang2018:DeepGraphs}.
That was possible because of the ubiquity of graphs and the solid capacity of neural networks in excelling learning representations from raw data.
The joining of graphs and computing techniques enables us to bring light to characteristics of interest that are not obvious for human inspections based on simple reading, or even for naïve algorithms.
Research on this topic has promoted substantial engagement due to the use of extensive and convoluted networks, and because such structures convey non-trivial patterns based on ingenious algorithms.
Although neural networks are fully-differentiable end-to-end computational graphs, the end-to-end graph-structure processing with neural networks is still in the early stages of discussion, such that research on graph-inspired neural networks is currently under the spotlight~\cite{Xu2019:Powerful}.
These approaches still do not portray complex systems as complex networks can do.
However, graph-inspired models bring the ability to analyze the graph topology by navigating inside its neighborhood through linear algebra operations on adjacency matrices.

Consequently, this thesis work delivers results from classic graph techniques to cutting-edge graph-inspired deep-learning methods.
Our contributions leveraged statistics, machine learning, and artificial neural networks.
Through such techniques, we improved the analysis, modeling, and organizational understanding of different human phenomena inherent to graphs that arise from the spatial interaction of epidemic propagation processes, complex networks of human migration, and geometric graphs derived from the structure of cities.
Containing three research fronts, the spinal cord of this work lies on a \textit{non-trivial data-driven modeling using artificial intelligence}.
We explored a broad domain of applications set by human phenomena emerging from social interaction observed in different granularities, such as between individuals, communities, and cities.
The convergent thematic of the thesis originated the following hypothesis:
\hfill%
\begin{adjustwidth}{.125\linewidth}{0\linewidth}
    \noindent\textit{\textbf{Thesis Hypothesis ---}}
    \textit{``The analytic processing by means of complex networks and graph metrics combined with artificial intelligence can expand the comprehension and, consequently, the capacity of modeling and forecasting human phenomena, providing us with information for acting on complex processes (\ie, pandemics progression over time), dynamic social interactions (\ie, pendular migration between cities), and on the network topology of cities (\ie, street networks from maps).''}~\cite{Spadon2021:Thesis}
\end{adjustwidth}

As a result, we obtained three main contributions, but the contributions are not limited to those three.
Firstly, we devised a graph-inspired learning-representation layer and neural network architecture for modeling spatial and temporal dynamic processes over different granularities \cite{Spadon2021:ReGENN}.
We formulated applications for the COVID-19 pandemic, weather forecasting problems, and patient monitoring in intensive care units.
Secondly, we contributed with a novel methodology based on supervised machine learning classification and regression for link prediction on graphs of human mobility and migration between all the cities of Brazil \cite{Spadon2019:Commuters}.
We employed population censuses and urban indicators collected in 2010.
Lastly, we produced a distance-based technique for tracking inconsistent urban structures, which are vertices in regions of poor vehicle mobility regarding points of interest, including hospitals, police stations, and schools \cite{Spadon2018:MultiScale}.
We used geometric graphs as intermediate city representations where streets are edges and intersections define the nodes.
In the course of the thesis, we contend that the results mark strong evidence that advances have been made to challenging and relevant problems related to human phenomena, proving the central hypothesis.
The thesis work was assembled as a collection of articles whose contributions are subsequently presented in descending chronological order.

\section*{Summary of Contributions}

\begin{figure}[h]
    \centering%
    \includegraphics[width=.9\linewidth]{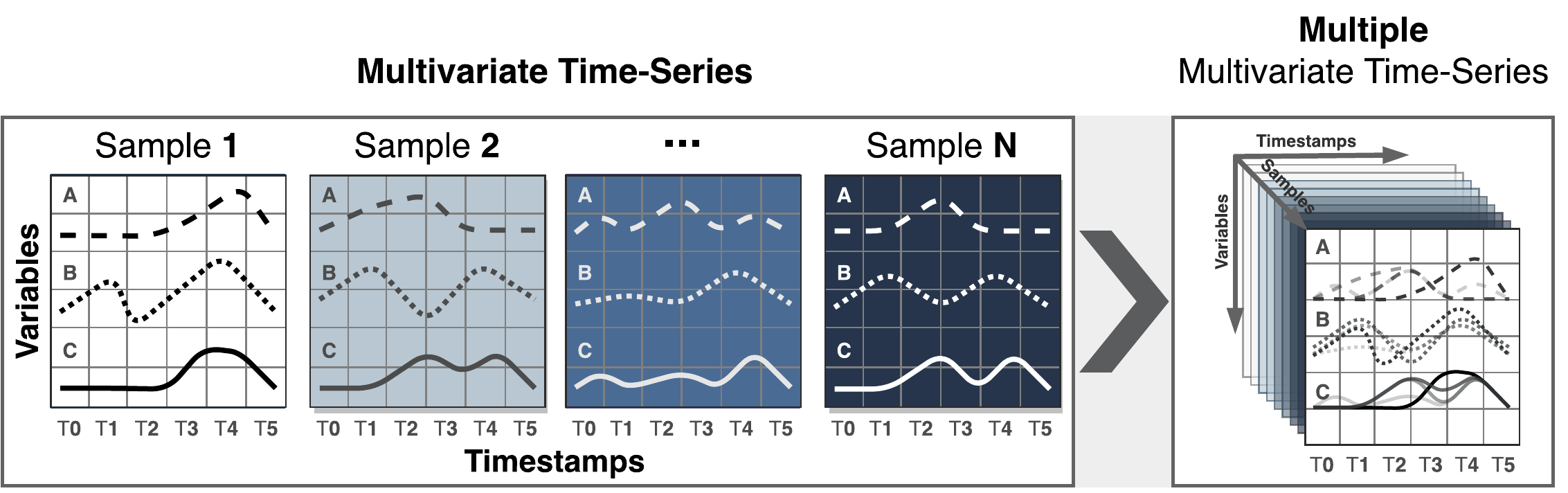}
    \caption{Problem definition and time-series data organization.}
    {\hfill\footnotesize{\bf Source:} Reproduced from~\cite{Spadon2021:ReGENN}.\hfill}
    \label{fig:6}
\end{figure}

\noindent\textbf{\textit{Dynamic Processes Modeling in Time.}}
Our first contribution is based on a graph-inspired learning-representation layer and neural network architecture.
Our technique is meant for modeling higher-dimensional time series by looking at multiple independent time series and all their related variables to leverage temporal patterns existing between different, yet correlated, data (see Figure~\ref{fig:6}).
As part of the results, we set a novel problem paradigm based on \textit{Multiple Multivariate Time Series}, which are stacked multivariate time series with the same variables observed during identical timestamps registered synchronously for various samples.
For example, when monitoring an endangered species, one requires understanding its habitat and simultaneously taking direct and indirect predators into account.
Such data arrangement yields an additional data dimension that can be understood as a multivariate sample of a higher-dimensional time series forecasting problem.
When working on such a class of problems, traditional neural networks and classical algorithms perform as ensembles by focusing on a single dimension of the data at a time, an approach that limits the information shared about different yet related data.

\begin{figure}[b]
    \centering\vspace{-.5cm}  
    \includegraphics[width=.9\linewidth]{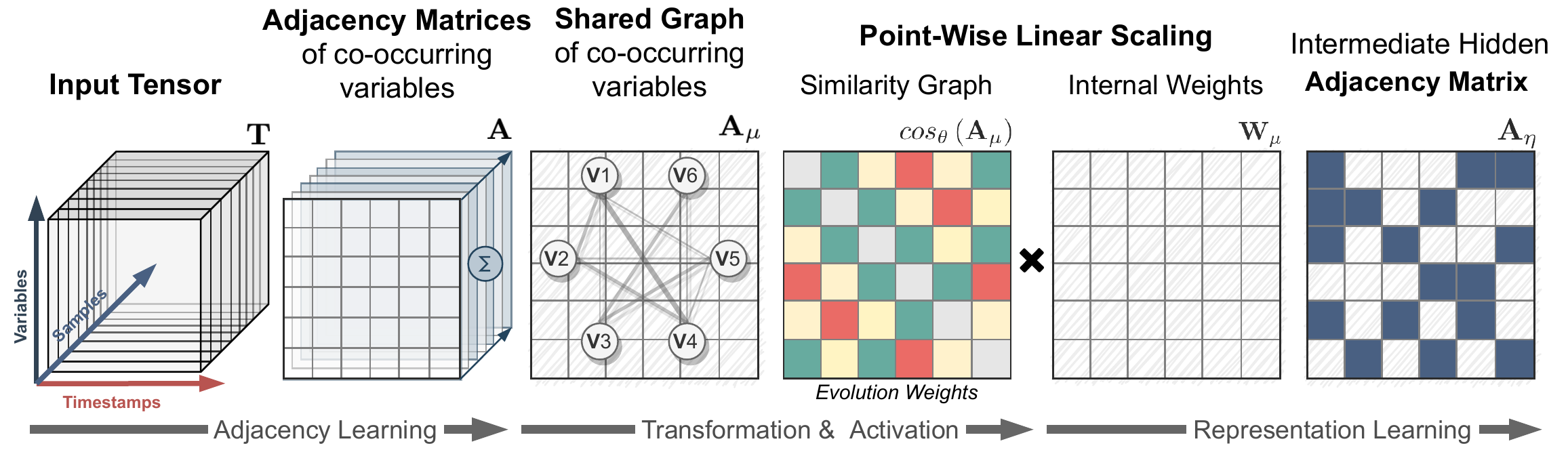}
    \caption{Graph Soft Evolution layer functioning.}
    {\hfill\footnotesize{\bf Source:} Reproduced from~\cite{Spadon2021:ReGENN}.\hfill}
    \label{fig:7}
\end{figure}

\begin{figure}[h]
    \centering
    \includegraphics[width=.9\linewidth]{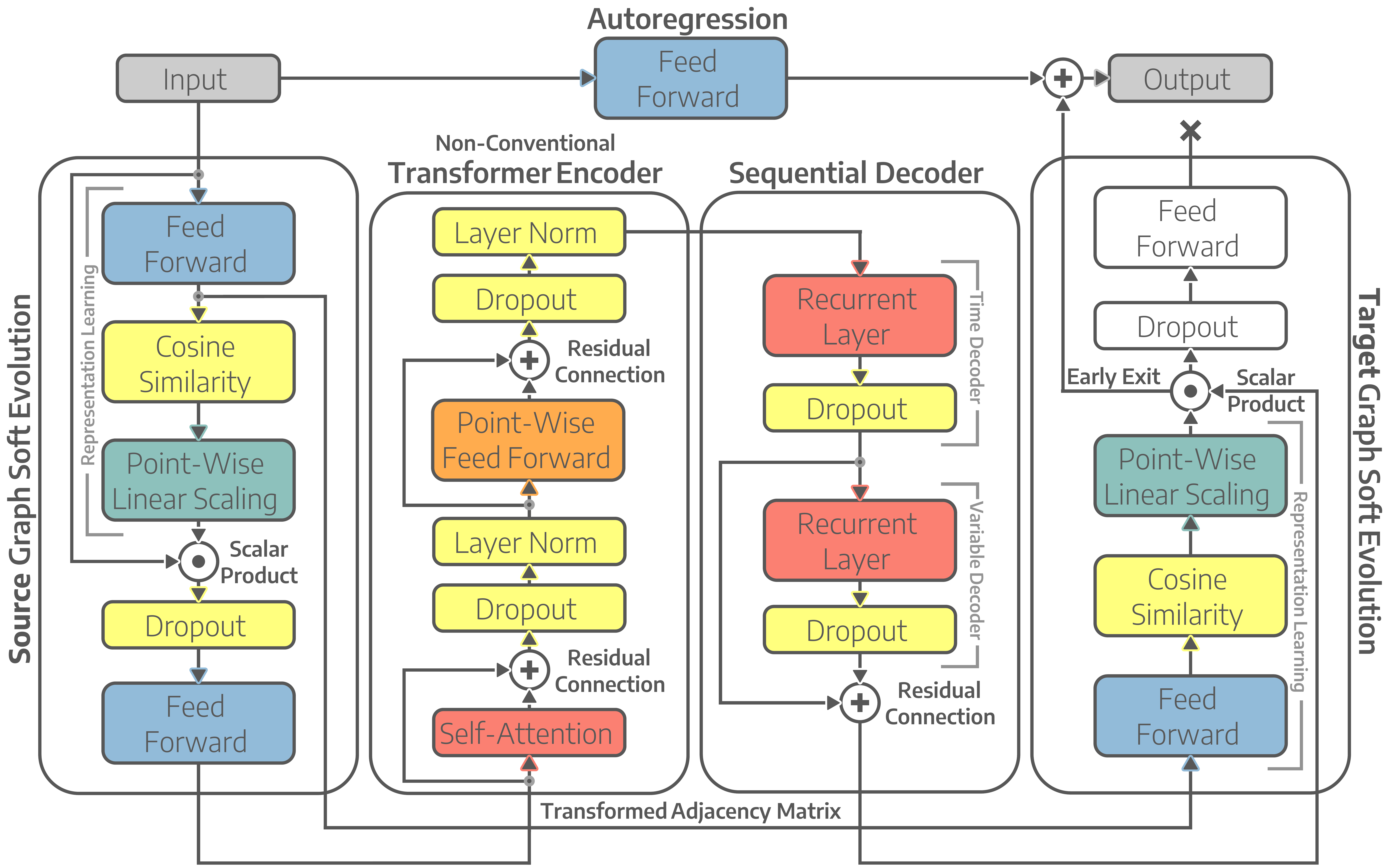}
    \caption{Recurrent Graph Evolution Neural Network.}
    {\hfill\footnotesize{\bf Source:} Reproduced from~\cite{Spadon2021:ReGENN}.\hfill}
    \label{fig:8}
\end{figure}

Thereby, we benefit from multiple multivariate time series to propose a new layer architecture and neural network, which are named, respectively, Graph Soft Evolution (GSE) and Recurrent Graph Evolution Neural Network (\ReGENN).
The GSE is a learning-representation layer that learns a shared graph from the training samples between the mutual variables existing in the time series, which is later converted into a similarity graph that will resemble the forecasted data (see Figure~\ref{fig:7}).
The GSE layer is part of \ReGENN, a graph-inspired time-aware auto-encoder with linear and non-linear pipelines working in parallel to jointly deliver predictions for the future using observations from the past (see Figure~\ref{fig:8}).
The linear pipeline of \ReGENN~stands for an Autoregression pipeline inspired by highway neural networks, while the non-linear is a transformer-based auto-encoder that operates with a pair of GSE layers, one at the beginning of the pipeline and the other at the end.
The first GSE layer learns a shared graph from the training data, creating a representation that describes the data from several time series and timestamps.
On the other hand, the second GSE layer re-learns the graph after the decoding, representing a graph of interaction potentially existing in the target data.

Our results are based on a full-spectrum benchmark of 50 algorithms ranging from classic time-series and machine learning algorithms to cutting-edge neural-networks-based ones.
Among those were the state-of-the-art multivariate time-series forecasting algorithms, such as the {\sc MLCNN} from {\sc AAAI'20}~\cite{Cheng2020:MLCNN}, {\sc DSANet} from {\sc CIKM'19}~\cite{Huang2019:DSANet}, and {\sc LSTNet} from {\sc SIGIR'18}~\cite{Lai2018:LSTNet}, and, even facing such stellar lineup, \ReGENN~surpassed all the tests reaching state-of-the-art positioning.
We experimented on the COVID-19\footnote{~Available at \url{https://github.com/CSSEGISandData}.}, Brazilian Weather\footnote{~Available at \url{http://bancodedados.cptec.inpe.br}.}, and 2012 PhysioNet Computing in Cardiology\footnote{~Available at \url{https://physionet.org/content/challenge-2012/1.0.0}.} datasets to assess the performance.
In the task of epidemiology modeling on the COVID-19 dataset, we observed up to 64.87\% improvement.
For the climate forecasting task on the Brazilian Weather dataset, we had up to 11.96\% improvement.
In patient-monitoring tasks on intensive care units (ICUs) on the PhysioNet dataset, we improved up to 7.33\%.
The results were accepted for publication at the \textit{IEEE Transactions on Pattern Analysis and Machine Intelligence} in a future issue of the journal and are currently listed on the journal's website as a peer-reviewed pre-print.
Because predicting events is a basic premise related to decision-making in urban planning, consumer behavior modeling, market analysis, and others, our novel layer and network architectures can be comprehensively beneficial to many applications in different areas such as seismic inversion and vessel mobility forecasting~\cite{CEMEAI2021:SeismicInversion, Spadon2022:AIS}.

\begin{wrapfigure}{r}{9cm}
    \includegraphics[width=\linewidth]{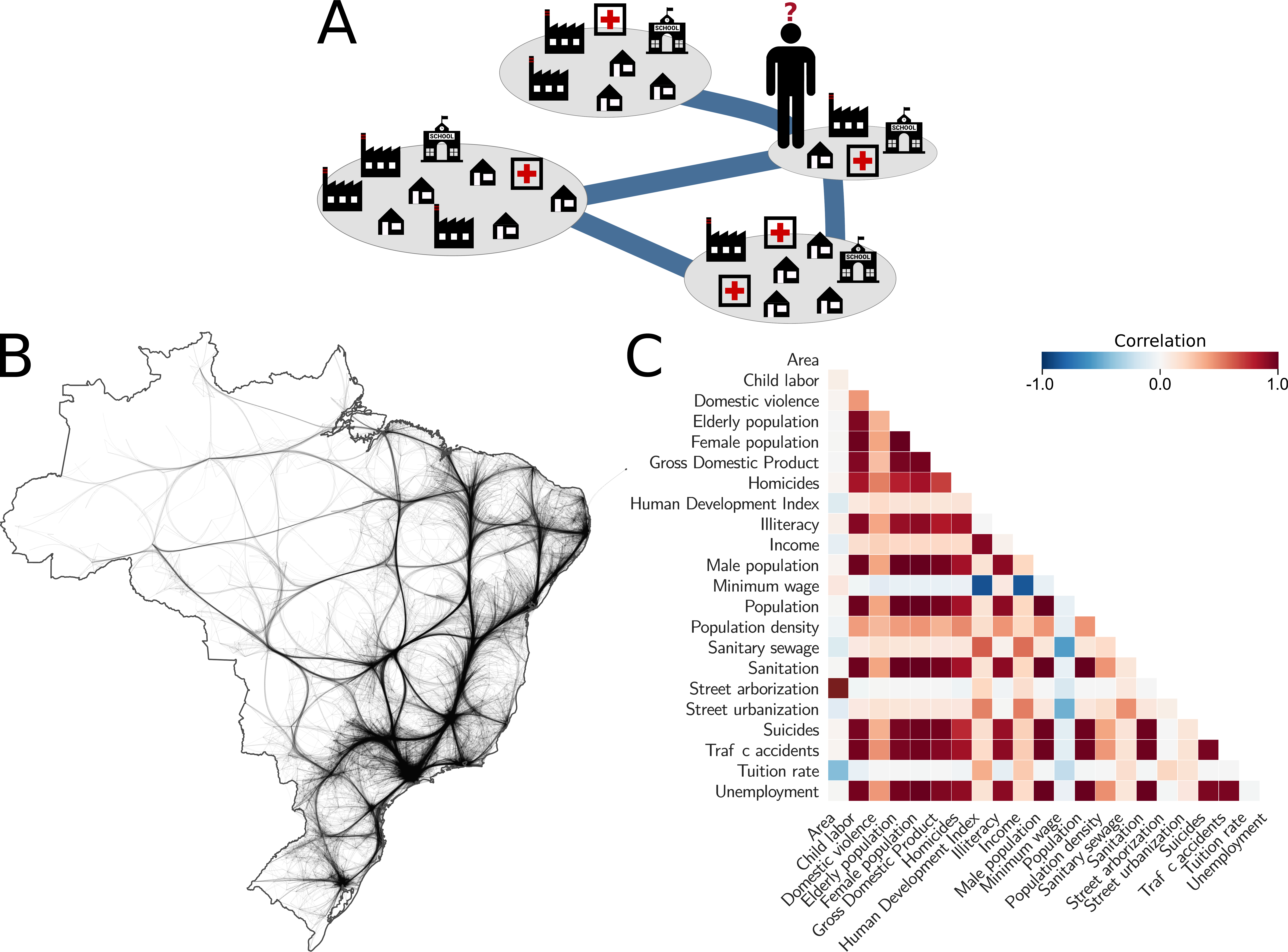}
    \captionof{figure}{Problem definition and feature analysis.}
    {\hfill\footnotesize{\bf Source:} Reproduced from~\cite{Spadon2019:Commuters}.\hfill}
    \vspace{.15cm}  
    \label{fig:3}
\end{wrapfigure}

\noindent\textit{\textbf{Human Mobility Forecasting.}}
Our second contribution advanced with a machine learning modeling methodology able to reconstruct the Brazilian inter-city commuters network using urban indicators on population census data (see Figure~\ref{fig:3}).
\begin{figure}[b!]
    \centering\vspace{-.5cm}  
    \includegraphics[width=.95\linewidth]{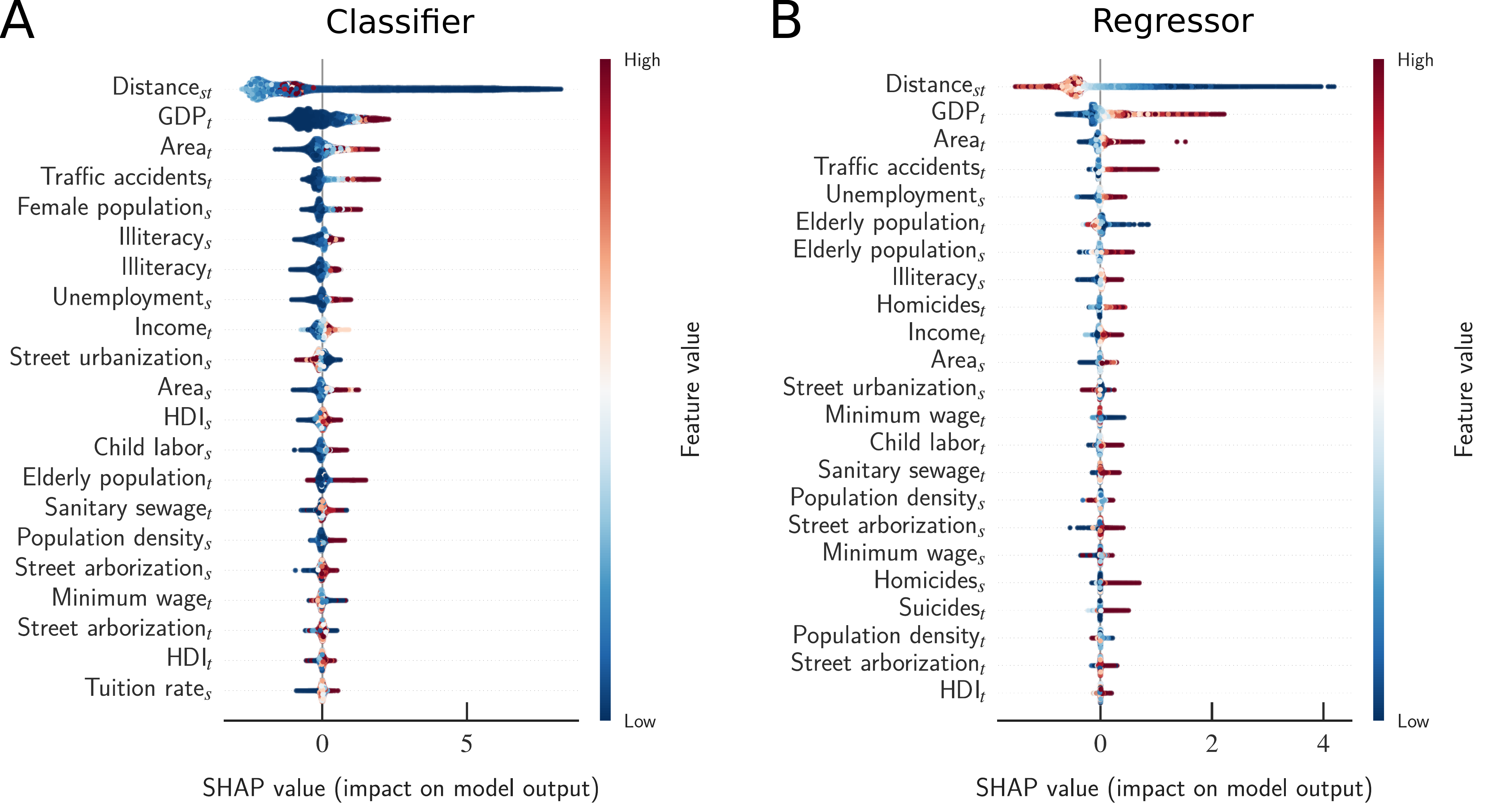}
    \caption{SHAP values analysis for feature importance assessment.}
    {\hfill\footnotesize{\bf Source:} Reproduced from~\cite{Spadon2019:Commuters}.\hfill}
    \label{fig:5}
\end{figure}
We provided an approach that could correctly describe the \textit{pendular migration} phenomena, representing the case of workers living in one city but working in a different city, thus, daily commuting between the two sites.
The proposal is based on the fact that the related literature has models predominantly based on the populational size and distance between the interacting cities.
The central literature on this topic was published by Nature's main journal and describes the Gravitation model~\cite{Simini2012:UniversalModel}, which, based on \textit{Newton's Law of Universal Gravitation}, considers the migration between cities as a function of distance and population sizes.
Another related work refers to the Radiation model~\cite{Ren2014:CommuterFlows} published in Nature Communications, which describes migration as a radiation and absorption phenomenon considering the population and distance between the interacting cities and the population of others up to a certain distance threshold.
Our research shows that both models struggle to accurately describe the commuters network due to not accounting for variables that have the potential to explain the reason behind migration phenomena.
As a result, we propose a model that, based on supervised machine learning, can predict the existence of migration (\ie, links) between two cities (\ie, nodes), leveraging many other variables that describe the quality of life and work in cities.
We also brought interpretability to the modeling proposal, highlighting the factors impacting commuter fluxes between cities.
At the same time, we determined the reasons that lead people to live in cities other than where they work, which are shown to be linked to the cities' quality of life and economic potential.

\begin{wrapfigure}{r}{10cm}
    \includegraphics[width=\linewidth]{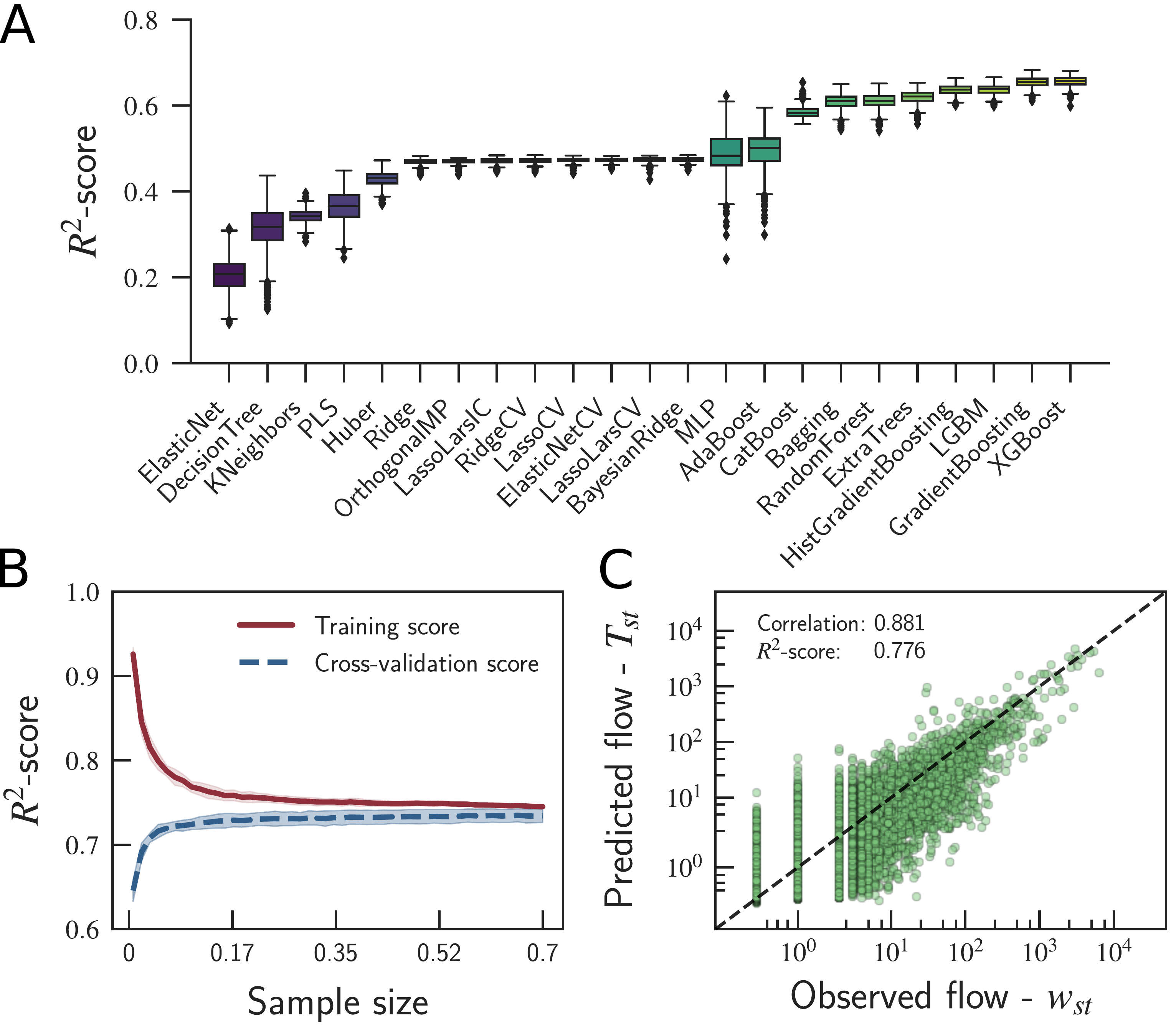}
    \captionof{figure}{Migration modeling with machine learning.}
    {\hfill\footnotesize{\bf Source:} Reproduced from~\cite{Spadon2019:Commuters}.\hfill}
    \label{fig:4}
\end{wrapfigure}
These results were published on \textit{Scientific Reports}~\cite{Spadon2019:Commuters}, in which we detailed how we put to the 78 test algorithms in tasks of classification and regression through statistical bootstrapping, learning curve analysis, and cross-validation (see Figure~\ref{fig:4}) while aiming to reveal the most prominent modeling approach in light of open-source urban indicators\footnote{~Available at \url{https://www.ibge.gov.br}.}.
We used classification to predict whether there is a migration between two cities and regression to forecast its intensity whenever the migration exists.
We applied bootstrapping and cross-validation to select the most statistically significant technique and learning-curve analyses to assess training generalization and overfitting.
Such an approach revealed that gradient-based algorithms could reconstruct the commuters' network with 90.4\% accuracy while describing 77.6\% of the variance observed in the number of people flowing between cities.
The essential features required to rebuild the commuters network using SHAP values analysis reinforced the fact that distance plays a vital role in migration, but that other indicators are also essential (see Figure~\ref{fig:5}).
This is the case of the features attracting workers to commute, such as high Gross Domestic Product (GDP) and a low unemployment~rate.

Modeling migrations allow for a better understanding of the population's organization and wealth distribution, being meaningful for assessing public policies regarding the regional economy and territorial planning.
Moreover, in the absence of population censuses data, such as the case of the \textit{Brazilian Population Censuses of 2020} that was postponed due to budget issues, such a model can provide estimates on the scarcity of factual data so public policies can be directed where needed.
Additionally, migration has a significant potential to explain other human-related phenomena, such as the case of crime that can be connected to the lack of jobs in particular cities.
Because the interaction between entities can be broadly observed in nature, expanding our knowledge about society and ourselves, one can extrapolate our findings to other systems, such as international trading, the spread of epidemics, social networks interactions, and food chains.

\noindent\textit{\textbf{Street Network Analysis.}}
Our third contribution advances with a set-theory-based and distance-driven pattern-discovery algorithmic technique for detecting vertices that lack access from/to points of interest in a city due to being in regions of poor vehicle mobility.
This technique combines the euclidean distance with the shortest path distance to find inefficient paths and the vertices that, by absorbing such inefficiency, turn into city regions of low mobility indices.
Such a proposal has roots in the concept of Accessibility~\cite{Travencolo2008:Accessibility}, defined through entropy, which is capable of assessing a city as a geometric graph by means of the accessibility of virtual edges.

\begin{wrapfigure}{r}{8cm}
    \includegraphics[width=\linewidth]{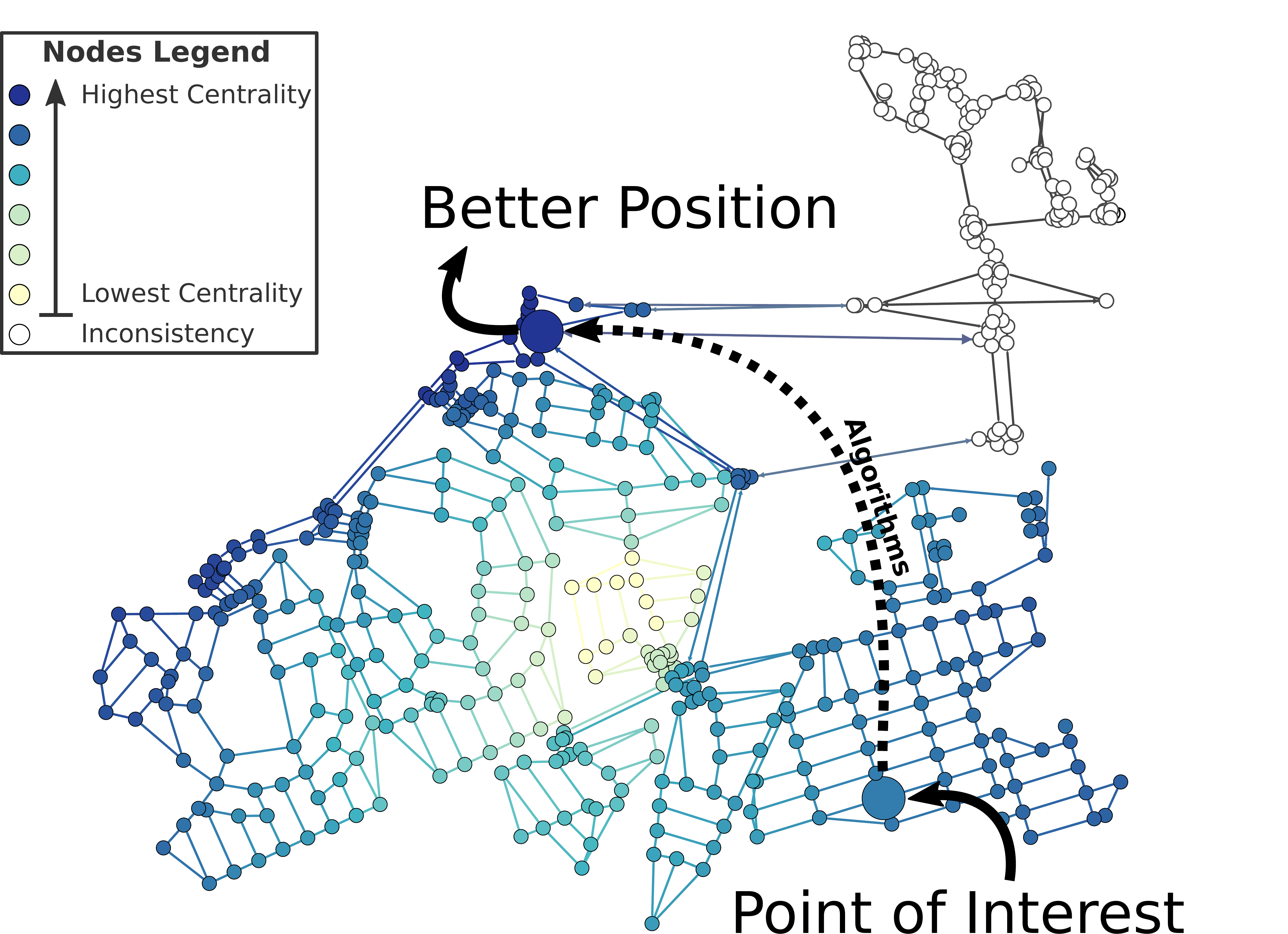}
    \captionof{figure}{Induced graph optimization workflow.}
    {\hfill\footnotesize{\bf Source:} Reproduced from~\cite{Spadon2018:MultiScale}.\hfill}
    \label{fig:1}
\end{wrapfigure}

Instead, our proposal considers the city a system of many parts, focusing on the vertices and edges while analyzing their inherent paths.
Consequently, we aimed at refining knowledge about cities' mobility based on user-given points of interest to improve interventions in the urban plan (see Figure~\ref{fig:1}).
The improvement, in this case, is measured by the number of inconsistent structures found in the geometric graph.
The near-optimal solution is a heuristically created graph with points of interest placed on locations that provide democratic access to most of the vertices in a city, as verified by centrality indicators.

\begin{wrapfigure}{l}{8cm}
    \includegraphics[width=\linewidth]{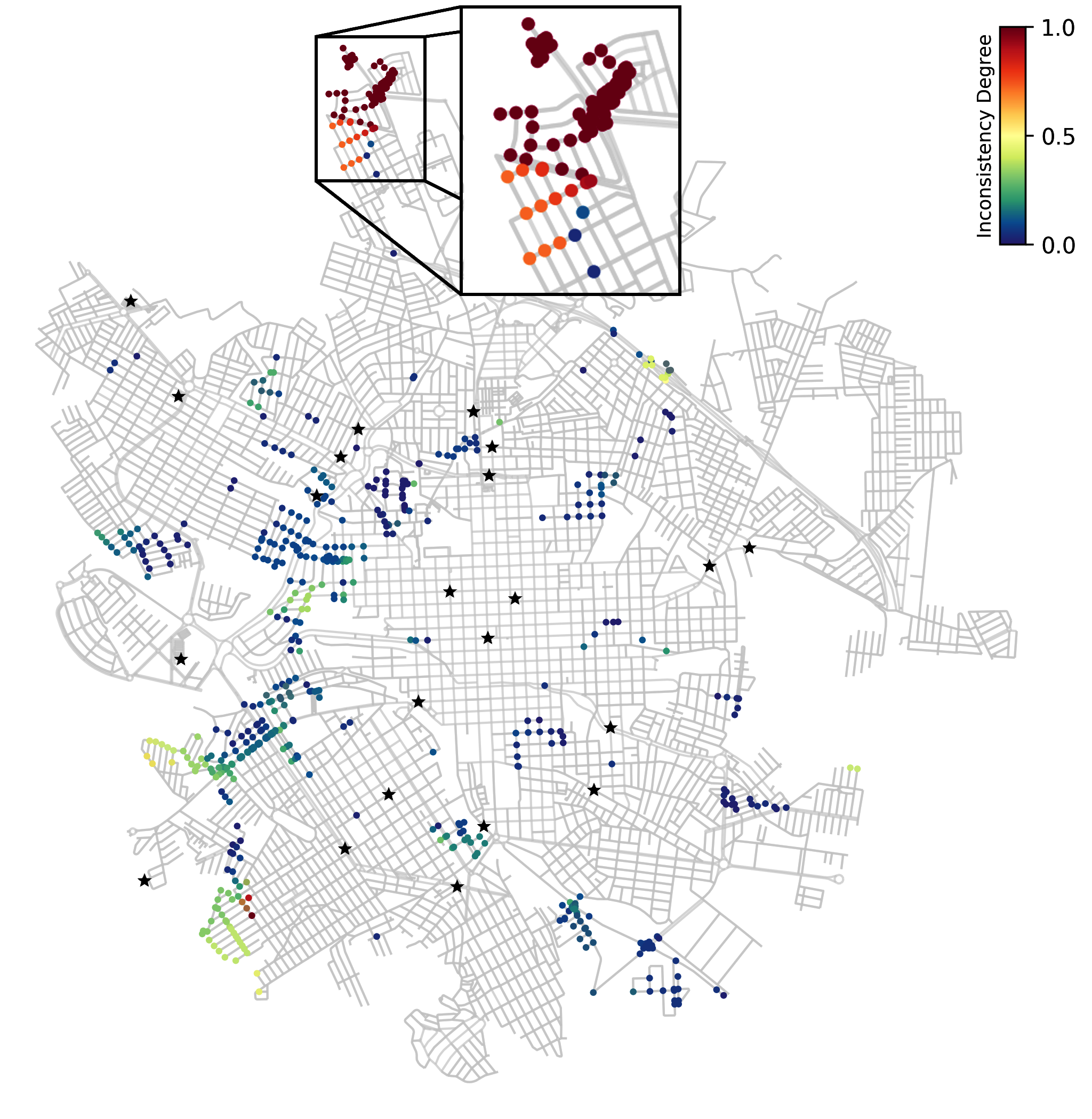}
    \captionof{figure}{Inconsistency degree of vertices.}
    {\hfill\footnotesize{\bf Source:} Reproduced from~\cite{Spadon2018:MultiScale}.\hfill}
    \label{fig:2}
\end{wrapfigure}

The proposed technique was based on open-source data\footnote{Available at~\url{https://www.openstreetmap.org}.} and conceived to provide support for decision-making related to resource location-allocation problems.
It is available to the public\footnote{Available at~\url{https://www.github.com/gabrielspadon}.} and can be used, for instance, in the initial design and early stages of the project of a city or neighborhood when considering building one or more points of interest (see Figure~\ref{fig:2}).
Our methodology proved to find better placements for points of interest while enhancing access indicators to most vertices in a city represented as a geometric graph.
As a result, we contributed with a concept based on intrinsic problems to urban structures, algorithms to track and heuristically reduce inconsistent vertices in geometric graphs, and case studies showing how our toolset and algorithms can aid urban planners.
The achieved results systematically treat a recurrent issue of broad interest in cities.
Nevertheless, our toolset is suitable to model multiple scenarios in which the vertices and edges positioning must be taken into account.
This case refers, for instance, to computer networks when adding or reallocating a switch or router; to the topological design of electronic circuits when it is possible to save on tracks by redistributing some components; or, to supply chains when it is possible to improve profit by better-distributing certain products across specific locations in the warehouse network.

The contributions in this section were included as {\it future~works} of an anomaly detection concept for urban agglomerations~\cite{Spadon2017:UrbanInconsistencies}, which was proposed by the Ph.D.
candidate in his Master's dissertation~\cite{Spadon2017:Dissertation}.
Such~proposal evolved into a more extensive refined set of techniques~\cite{Spadon2018:InconsistentStructures} and side contributions~\cite{Spadon2018:TopologicalCharacterization, Spadon2018:CityTouring}, becoming one of the best papers of the \textit{18th International Conference on Computational Science}, Wuxi -- China.
We were invited to extend our proposal~\cite{Spadon2018:MultiScale} and publish it in the \textit{Journal of Computational Science}, in which we went further in the analysis of different points of interest in urban structures while exploring the concept of \textit{walkability}, adapting the previous technique by incorporating walking-paths as part of the mobility processes in cities, showing potential to improve the walking-mobility.

\section*{Final Remarks}
The thesis contributes with methodologies for analyzing spatial and temporal data from the perspective of computer science applied to \textit{(i) Dynamic Processes Modeling in Time}, \textit{(ii) Human Mobility Forecasting}, and \textit{(iii) Street Network Analysis}.
We contribute to problems that manifest on multiple scales, from the simultaneous analysis of whole countries expressed as time series to the topological issues existing within street networks.

We accomplished advances in a broad range of domains, with applications in real-world scenarios of medicine, weather, human mobility, and urban organization. The employed techniques encompassed the vast universe of artificial intelligence and machine learning, including artificial neural networks, prediction/regression supervised learning and statistic-based algorithms. In common, we used graphs to represent the complex networks that underlie the problems we faced. As evidenced by our literary production,
we brought forth knowledge sustained by rigorous formalisms, intense experimentation, and innovative ideas in consonance with the state of the art in Computer Science.
In conclusion, the thesis not only advances with computational techniques but also with theoretical principles able to propel knowledge about human-related phenomena. Our results were peer-reviewed and published in top-tier computer science journals, while other thesis-related contributions were published as book chapters and in conference proceedings.

\section*{Acknowledgments}
The thesis work and the products derived from it were supported (directly or indirectly) by the Coordena\c{c}\~ao de Aperfei\c{c}oamento de Pessoal de N\'ivel Superior -- Brazil (CAPES) -- Finance Code 001; Funda\c{c}\~ao de Amparo \`a Pesquisa do Estado de S\~ao Paulo (FAPESP), through grants 2013/07375-0, 2014/25337-0, 2016/02557-0, 2016/16987-7, 2016/17078-0, 2017/08376-0, 2018/17620-5, 2019/04461-9, and 2020/07200-9; Conselho Nacional de Desenvolvimento Cient\'ifico e Tecnol\'ogico (CNPq) through grants 167967/2017-7, 303694/2015-7, 305580/2017-5, 404870/2016-3, and 406550/2018-2; National Science Foundation awards IIS-1838042, IIS-2014438, and PPoSS-2028839; and, the National Institute of Health awards NIH R01 1R01NS107291-01, and R56HL138415.

\bibliographystyle{sbc}
\bibliography{references}

\section*{Authors Biography}

\vspace{.2cm}\noindent%
\begin{wrapfigure}{l}{2.5cm}
    \includegraphics[width=1in, height=1.5in, clip, keepaspectratio]{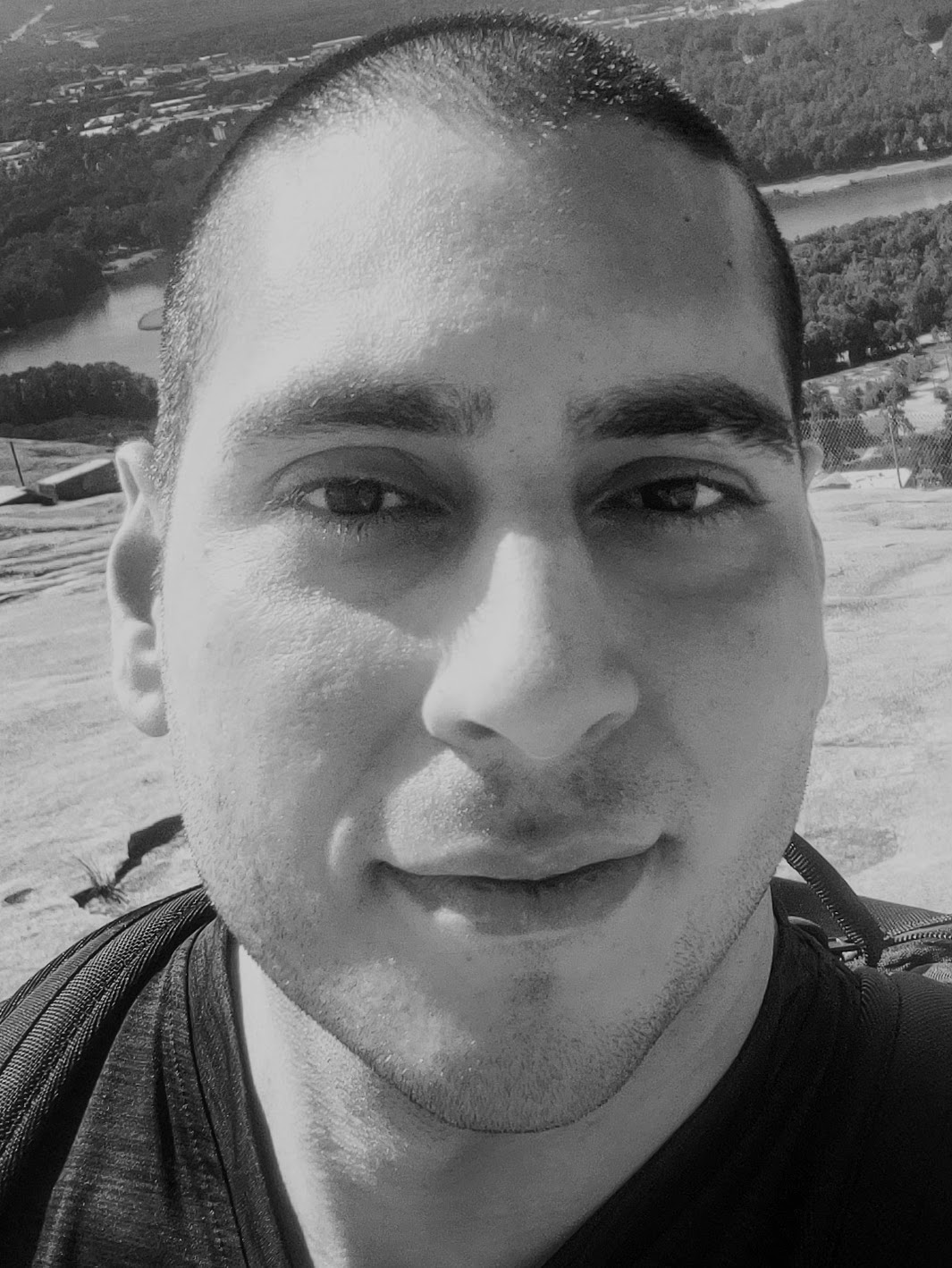}
\end{wrapfigure}
\noindent%
{\bf Gabriel Spadon} is currently a postdoctoral fellow at Dalhousie University, Canada, working on projects related to vessel mobility and underwater acoustics to architect neural networks for improving ocean awareness and monitoring capabilities. He has a Ph.D. (with honors) in Computer Science at the University of Sao Paulo, Brazil, part of which was carried out at the Georgia Institute of Technology, USA. Spadon has worked intensively on network science and artificial intelligence during the last few years. He has authored (and co-authored) several research articles on knowledge discovery through complex networks and data mining. His current research interests include neural-inspired models, graph-based learning, and complex networks.

\vspace{.2cm}\noindent%
\begin{wrapfigure}{r}{2.5cm}
    \includegraphics[width=1.05in, height=1.75in, clip, keepaspectratio]{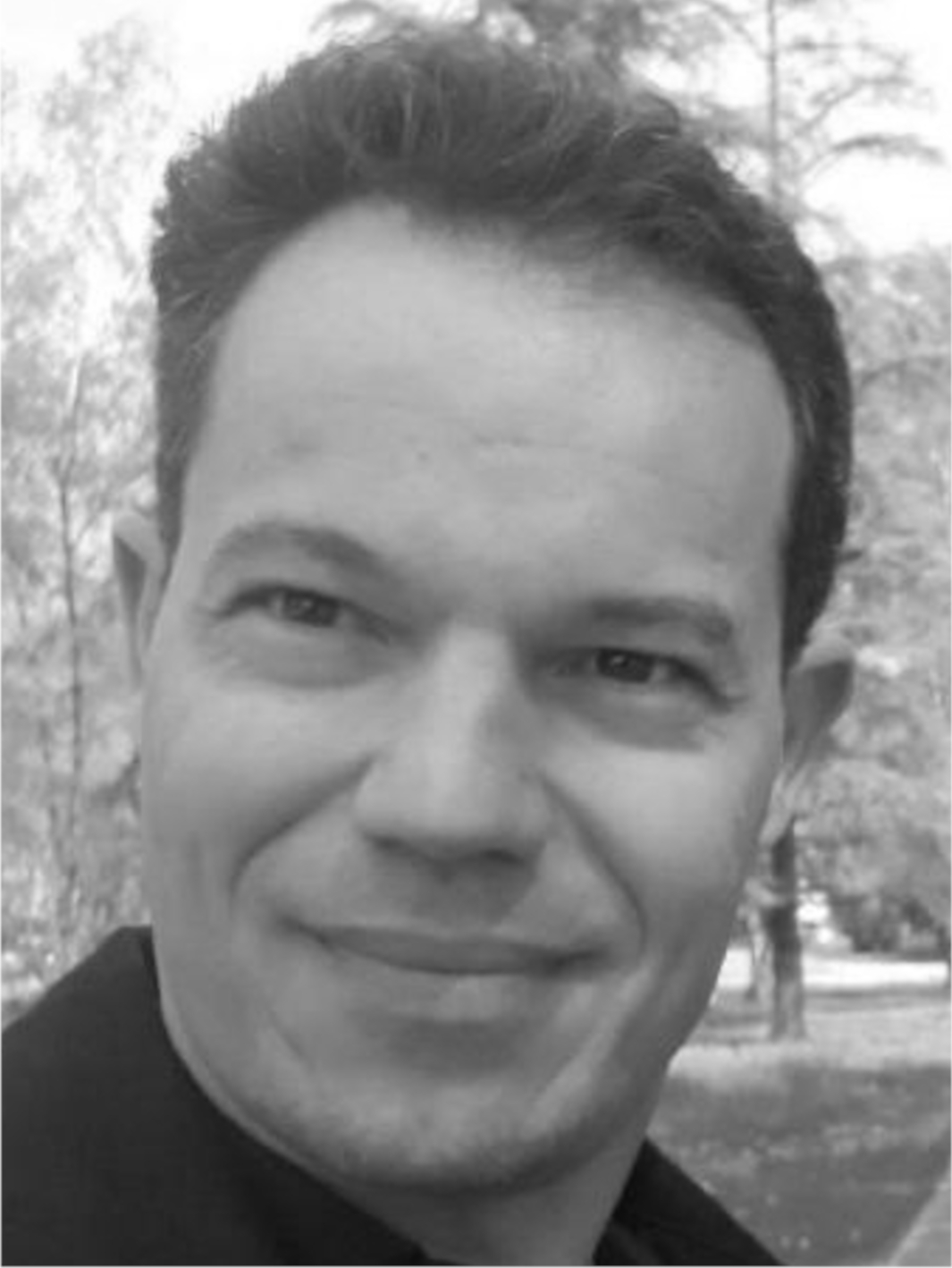}
\end{wrapfigure}
\noindent%
{\bf Jose F. Rodrigues-Jr} received the Ph.D. from the University of Sao Paulo, Brazil, part of which was carried out at Carnegie Mellon University, USA, in 2007. He is currently an associate professor at the University of Sao Paulo, Brazil. He is a regular reviewer and author in his research field, which includes data science, machine learning, content-based data retrieval, visualization, and the application of such techniques in the medic, agriculture, and e-learning domains, contributing to publications in major journals and conferences in his area of expertise.

\end{document}